\documentclass[10pt,twocolumn,letterpaper]{article}

\usepackage[pagenumbers]{wacv} 

\usepackage{makecell}   
\usepackage{tabularx}  
\usepackage{booktabs}
\usepackage{multirow}
\newcolumntype{R}{>{\raggedleft\arraybackslash}X}

\definecolor{wacvblue}{rgb}{0.21,0.49,0.74}
\usepackage[pagebackref,breaklinks,colorlinks,allcolors=wacvblue]{hyperref}

\def\wacvPaperID{*****} 
\def\confName{WACV}
\def\confYear{2027}

\title{Auditing Latent-Space Monitors for Autonomous Driving}

\author{Nikhil Kamalkumar Advani\\
Independent Researcher\\
{\tt\small nik.advani16@gmail.com}
\and
Vishwajeet Shivaji Hogale\\
Noertheastern University\\
{\tt\small Hogale.v@northeastern.edu}
\and
Saurav Kumar\\
Bosch North America\\
{\tt\small Saurav.kdeo@gmail.com}
}

\begin{document}
\maketitle
\begin{abstract}
Runtime failure monitors can use a model's internal representations to
anticipate failures. We audit this monitoring strategy across two
autonomous-driving tasks: online vectorized map generation with LaneSegNet and
end-to-end planning with VAD. We find that frame-level errors are predictable at
inference in both tasks. For LaneSegNet, a supervised latent probe reaches Area Under the Receiver Operating Characteristic curve (AUROC) $0.780$ for high Chamfer error; to our knowledge, this is
the first post-hoc frame-level failure monitor for online vectorized map
generation. For VAD, a supervised planning-latent probe reaches AUROC $0.868$
for mean-ADE failure.

Our audit shows that internal access is not necessary for strong failure prediction. A monitor using only LaneSegNet's
prediction outputs reaches AUROC $0.825$, while for VAD, ego state, driving
command, and the planner's predicted trajectory reach $0.924$ on the same
mean-ADE endpoint. Adding latent features to either baseline yields no statistically resolved improvement. This observation persists across a broad suite of planning failure endpoints, including endpoints whose labels depend on geometry unavailable to the non-latent baseline. Thus, predicting failure from an internal representation does not establish that the representation provides useful information beyond observable inputs and outputs. We propose an evaluation protocol for testing the incremental value of latent access and release our per frame failure endpoint labels.
\end{abstract}    
\section{Introduction}
\label{sec:introduction}

Runtime failure monitoring asks when an autonomous system's perception or
plans are likely to be unreliable, so that they can be rejected, deferred, or
given additional scrutiny. A natural place to look is inside the model itself.
Intermediate representations encode the scene before it is compressed into a
final prediction or action, and may expose difficulty that the output does not
reveal. Accordingly, prior work has used learned representations for
out-of-distribution detection, uncertainty estimation, and supervised failure
prediction across the driving
stack~\cite{daftry2016introspective,wiederer2023joint,xiong2025riskmonitor}.

We audit this practice on two structurally different autonomous driving tasks
and initially find strong evidence in its favor. For online vectorized map
generation with LaneSegNet~\cite{li2024lanesegnet}, a supervised probe over
frozen image and BEV representations predicts high matched-lane
Chamfer error at Area Under the Receiver Operating Characteristic curve (AUROC) $0.780$. For end-to-end planning with
VAD~\cite{jiang2023vad}, a probe over the frozen latent planning representation
reaches AUROC $0.868$ for mean ADE greater than
$2\,\mathrm{m}$.

We then investigate whether internal access is actually necessary, and find that it is not.
For LaneSegNet, a monitor
using only prediction outputs reaches AUROC $0.825$, which is higher than the
latent-only monitor. For VAD, ego state, driving command, and the planner's own
predicted trajectory reach $0.924$ on the same ADE endpoint mentioned above. Adding the
corresponding latent representations back to either baseline yields no
statistically resolved improvement. Thus, the tested internal features provide
no demonstrated incremental predictive value once strong non-latent baselines
are included.

For VAD, since ADE is itself defined from the
planner's predicted trajectory, prediction from that trajectory is
potentially circular. We therefore repeat the audit on endpoints whose labels
require information unavailable to the non-latent monitor, including
agent proximity and road boundary contact.
Latent augmentation again yields no statistically resolved improvement.

These experiments expose a distinction that is easy to miss. A representation can have strong \emph{marginal} predictive value
for failure while providing little or no \emph{incremental} value once the
system's observable state and outputs are known. Predicting failure from a
latent representation therefore does not establish that internal access is
needed. We argue that latent space failure monitors should be evaluated against
strong output and state based baselines and tested directly for incremental
predictive value.

\noindent\textbf{Contributions:}
\begin{itemize}
    \item We develop and audit inference-time failure monitors for online vectorized map generation with LaneSegNet and end-to-end planning with VAD. For the former we introduce, to our knowledge, the first  post-hoc
    monitor that predicts frame-level failure defined by geometric error for a frozen
    online vectorized map generation model.

    \item We show that runtime-visible state and prediction output baselines outperform
    the headline latent monitors for LaneSegNet and VAD, while latent
    augmentation yields no statistically resolved gain.

    \item We propose an evaluation protocol that tests latent monitors against
    strong non-latent baselines and directly measures the incremental value of
    internal access, including on endpoints whose defining geometry is hidden
    from those baselines. 
    
    \item We release per-frame evaluation annotations for 61,432 samples across
    VAD and LaneSegNet including continuous error quantities,
    and decomposed endpoints spanning trajectory disagreement,
    agent collision and clearance, boundary contact, and lane-geometry quality, thereby
    enabling reproducible evaluation.
    
\end{itemize}
\section{Related Work}

\noindent\textbf{Monitoring perception from internal representations:}
Early introspective methods estimate perception failures from sensor
observations or learned representations~\cite{daftry2016introspective,kuhn2020introspective},
while loss-prediction heads regress model error from internal
features~\cite{yoo2019learning}. Rahman et al.\ bring this to per-frame object
detection, freezing a detector and training a classifier over pooled backbone
activations to flag frames whose per-frame mAP falls below a
threshold~\cite{rahman2020perframe}. Yatbaz et al.\ extend frame-level
activation-based introspection to automotive detectors and dataset
shift~\cite{yatbaz2024runtime}.

For online map generation, reliability has instead mainly been studied 
through uncertainty and robustness. Gu et al.\ augment online vectorized
mappers with point-wise uncertainty~\cite{gu2024producing}, COMap incorporates
Gaussian confidence estimates~\cite{huang2025comap}, and MapDiffusion learns a
distribution over plausible maps from which uncertainty can be
estimated~\cite{monninger2025mapdiffusion}. UIGenMap uses uncertainty-aware
structure injection for geographic generalization~\cite{liu2025uigenmap}. LanePerf~\cite{wu2025laneperf} estimates lane-detection performance, but operates on multi-frame
segments. Consequently, it can characterize
aggregate performance degradation over time. We instead formulate online vectorized map monitoring as a
per-frame failure-prediction problem, enabling individual predictions to be
rejected or flagged at inference time.

\noindent\textbf{Monitoring planning from internal representations:}
For trajectory prediction, Wiederer et al.\ ~\cite{wiederer2023joint} attach a latent Gaussian mixture
model and a supervised error-regression head to a frozen encoder, finding that
density detects novel scenarios while supervision better predicts
error. Farid et al.\ argue that failure
detection should target errors consequential to the downstream planner rather
than generic prediction error~\cite{farid2023task}, motivating the physically
grounded endpoints of Sec.~\ref{sec:vad}. For end-to-end planners,
RiskMonitor predicts collision-loss events in VAD and UniAD from internal
planning and road-user motion tokens~\cite{xiong2025riskmonitor}, making it the
closest precedent to our VAD latent-monitor experiments. REDOUBT combines
latent distribution-shift detection with output uncertainty for motion-planning
safety validation~\cite{wang2025redoubt}, while Argus monitors planned
trajectories and intervenes on detected hazards~\cite{wang2025argus}. These methods establish the usefulness of planning representations for failure prediction. We investigate whether those representations improve prediction after conditioning on ego state and the planner's generated trajectory.

\noindent\textbf{Failure prediction without internal representations:}
Kuhn et al.\ anticipate driving
disengagements from vehicle state and detected-object counts, treating the
system as a black box~\cite{kuhn2020blackbox}, and separately from sequences
of planned trajectories~\cite{kuhn2021trajectory}. On nuScenes, ego status is
unusually informative: AD-MLP attains competitive planning performance from
ego motion alone~\cite{zhai2023rethinking}, BEV-Planner shows that end-to-end
planners rely heavily on it~\cite{li2024egostatus}, and AdaptiveAD attributes
this shortcut in part to premature ego-status fusion~\cite{tang2026adaptivead}.
These results make ego state a necessary baseline when interpreting planning
latents.

Prior work establishes that internal representations can predict
failure, while observable state and output summaries can also carry
failure-relevant information. What remains underexplored is the conditional
question central to our study: after a strong non-latent monitor is given the
relevant observable information, does internal access add predictive value?
We test this directly by holding the endpoint and evaluation protocol fixed
and appending the latent representation to the strongest non-latent baseline.

\begin{table*}[t]
\centering
\caption{\textbf{Prediction of mean matched-lane Chamfer 
$>0.5$\,m for LaneSegNet}
$AP_{ls}^{*}$ denotes a frame-level analogue used only as a secondary
descriptive metric and is distinct from the official dataset-level
OpenLane-V2 $AP_{ls}$. Results are on the validation set.}
\label{tab:lanesegnet_results}

\footnotesize
\setlength{\tabcolsep}{2.7pt}

\begin{tabularx}{\textwidth}{@{}X l c c c cc cc@{}}
\toprule
Information available & Approach
& AP $\uparrow$ & AUROC $\uparrow$ & Coverage
& \multicolumn{2}{c}{Mean Chamfer (m)}
& \multicolumn{2}{c}{$AP_{ls}^{*}$} \\
\cmidrule(lr){6-7}
\cmidrule(lr){8-9}
& & & & & flagged & retained & flagged & retained \\
\midrule

All-camera FPN channel means (latent)
& PCA-128 + GMM NLL
& .545 & .636 & 49.6\%
& .552 & .454 & 0.305 & 0.405 \\

Road-focused per-camera image (latent)
& PCA-128 + GMM NLL
& .611 & .679 & 29.1\%
& .600 & .462 & .281 & .387 \\

BEV spatial $2{\times}2$ (latent)
& PCA-128 + GMM NLL
& .666 & .729 & 15.9\%
& .662 & .472 & .198 & .386 \\

\midrule

Road-focused image (latent)
& Supervised MLP
& .736 & .753 & 28.0\%
& .629 & .453 & .257 & .395 \\

BEV spatial $2{\times}2$ (latent)
& Supervised MLP
& .746 & .773 & 33.9\%
& .639 & .432 & .228 & .422 \\

Image + BEV (latent)
& Supervised MLP
& .761 & .780 & 32.3\%
& .639 & .437 & .240 & .411 \\

\midrule

Raw predicted lane set (non-latent)
& DeepSets-style encoder
& .787 & .809 & 33.7\%
& .654 & .425 & .241 & .415 \\

Engineered prediction output (non-latent)
& Supervised MLP
& \textbf{.792} & \textbf{.825} & 33.2\%
& .649 & .429 & .238 & .415 \\

Output + image + BEV (latent + non-latent)
& Supervised MLP
& .782 & .811 & 35.7\%
& .652 & .419 & .230 & .426 \\

\bottomrule
\end{tabularx}
\end{table*}

\section{Evaluation Protocol}
\label{sec:evaluation_protocol}

We audit latent failure monitors through three questions:

\noindent\textbf{Q1: Representation novelty}
We first ask whether model error increases when an internal representation
becomes atypical relative to training. Features are standardized, reduced by unwhitened PCA, and modeled with a training-fitted Gaussian mixture
model (GMM); negative log likelihood is the novelty score. This is a
failure-agnostic measure of \emph{representation novelty}, not a failure
probability.

\noindent\textbf{Q2: Supervised latent predictability}
Novelty may miss failure-relevant directions that remain common under the
marginal representation distribution. We therefore train lightweight
supervised probes over frozen internal representations to predict each task's
failure endpoint.

\noindent\textbf{Q3: Incremental latent value}
Finally, we test whether \emph{direct latent access} adds information beyond a
strong non-latent baseline. We compare that baseline with the same baseline
augmented by latent features, holding evaluation samples and the classifier and training protocol fixed. We repeat this comparison across alternative failure definitions,
including endpoints whose labels require scene information unavailable to the
non-latent monitor.

\noindent\textbf{Methodology:}
All fitting, feature transforms, model selection, calibration, and
operating-point selection use training data, grouping temporally related
LaneSegNet samples by source and VAD samples by scene. Validation data are not
used to fit any transform, monitor, calibration mapping, or threshold; reported
validation metrics are computed only after each monitor is fixed. We report
AUROC and average precision (AP), interpreting AP relative to endpoint
prevalence, and selective-risk curves where useful.

\section{Monitoring Online Vectorized Map Generation with LaneSegNet}
\label{sec:lanesegnet}

\begin{figure*}[t]
    \centering
    \includegraphics[width=\textwidth]{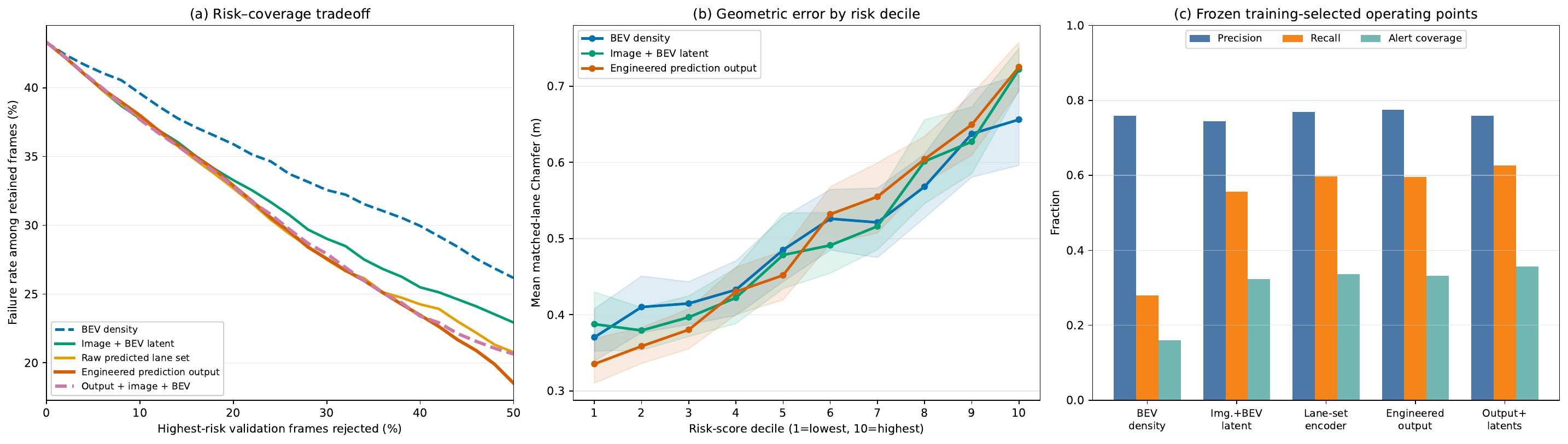}
    \caption{\textbf{LaneSegNet monitor hierarchy and incremental-value audit.}  All monitors predict mean matched-lane Chamfer \(>0.5\) m. Model selection and operating-point selection use training data only. (a) Failure rate among retained frames as increasingly high-risk frames are rejected. (b) Mean matched-lane Chamfer across risk-score deciles for representative density, supervised-latent, and prediction-output monitors; bands show 95\% source-bootstrap intervals. (c) Precision, recall, and alert coverage at frozen training-selected operating points. }
    \label{fig:lanesegnet_monitors}
\end{figure*}

We apply the audit of Sec.~\ref{sec:evaluation_protocol} to
LaneSegNet~\cite{li2024lanesegnet} on OpenLane-V2~\cite{wang2023openlanev2}. Prior work
modifies mapping models to expose uncertainty or
confidence~\cite{gu2024producing,huang2025comap,monninger2025mapdiffusion},
uses uncertainty to improve generalization~\cite{liu2025uigenmap}, or analyzes
robustness and failure modes offline~\cite{shan2026stability,hubbertz2026failuremodes}. To our
knowledge, this is the first frame-level monitor to predict failure defined by geometric
error for a frozen online vectorized map generation model at inference.

\subsection{Representations} 
\label{sec:lanesegnet_representations} 
\noindent\textbf{Image and BEV latents:} We initially considered three summaries of both image and BEV features: channel mean, channel mean and standard deviation, and spatial $2{\times}2$ pooling. For failure defined by the matched-lane geometric error, BEV spatial $2{\times}2$ produced the strongest density-based association among the BEV summaries and is therefore retained in the main comparison. The global image summaries were weaker, motivating a camera-aware representation that preserves each of LaneSegNet's seven camera streams and retains the lower $1{\times}2$ row of the spatial grid, where road evidence is concentrated. We use this road-focused representation for the final image density and supervised monitors. The latent-fusion monitor combines the road-focused image representation with BEV spatial $2{\times}2$ features. 

\subsection{Monitoring Approaches}
\label{sec:lanesegnet_methodology}

\noindent\textbf{Density-based representation novelty:}
Training source IDs are divided into a 90\% density-fit partition and a
disjoint 10\% calibration partition, preventing temporally related frames
from crossing the split. On the density-fit partition, features are
standardized, reduced by unwhitened PCA to 128 dimensions, and modeled with
a diagonal-covariance GMM. The component count is selected between 32 and 64
by training BIC, and negative log likelihood (NLL) provides the continuous
novelty score. For BEV, a single GMM is fitted to the spatial $2{\times}2$ representation.
For the road-focused image monitor, an independent scaler, PCA transform,
and GMM are fitted for each of the seven cameras. For both monitors, the disjoint
calibration partition is used to select a binary classification threshold by
maximizing Youden's $J$ for the failure endpoint.

\noindent\textbf{Supervised latent monitoring:} For supervised image monitoring, each road-focused camera block is independently standardized and reduced to 64 PCA coordinates, and the seven blocks are concatenated into 448 dimensions. BEV spatial $2{\times}2$ is reduced to 128 PCA coordinates. Image--BEV fusion therefore contains 576 coordinates. The latent heads use a $256$--$128$--$32$ MLP with LayerNorm, GELU activations, dropout $0.2$, AdamW, and inverse-prevalence positive weighting. Training uses five-fold source-grouped cross-fitting so that temporally
related frames remain together. Out-of-fold training predictions are used
to select the fit probability calibration, and choose the
binary classification threshold without using the validation set. The final
MLP is then trained from scratch on all finite-label training frames for the
selected number of epochs and evaluated once on validation.

\noindent\textbf{Model-output monitoring:} The structured lane set monitor uses the permutation-invariant DeepSets-style encoder described earlier. Separately, we train a supervised MLP on the engineered prediction output descriptor. We also evaluate a supervised MLP after appending the image and BEV latent features to the engineered output representation. The latter comparison asks directly whether access to the internal representations improves monitoring once LaneSegNet's observable prediction is already available.

\subsection{Failure Endpoint}
\label{sec:lanesegnet_endpoints}
Because OpenLane-V2 $AP_{ls}$ and mAP are confidence-ranked dataset-level
metrics, they do not directly provide the per-frame failure label required for
runtime monitoring. Moreover, applying an AP-style metric independently to individual frames yields a score that conflates missed detections, false positives, confidence ranking, and geometric accuracy, rather than directly measuring the severity of geometric error. Nevertheless, because $AP_{ls}$ is a standard evaluation
measure for lane-segment perception in online vectorized map
generation~\cite{li2024lanesegnet,abraham2026georeformer},
we report a frame-level analogue as a secondary descriptive metric ($AP_{ls}^{*}$). 

We define failure from the realized geometry of LaneSegNet's matched lane
predictions. Predictions are retained at confidence at least $0.30$ and within
$25\,\mathrm{m}$ of the ego vehicle; pedestrian crossings are excluded. Each eligible
ground-truth lane $i$ and prediction $j$ is scored using the relaxed
OpenLane-V2 3-D lane-segment distance
\begin{equation}
d_{ij}=r_i\frac{d_F(C_i,\hat C_j)+d_C(L_i,\hat L_j)+d_C(R_i,\hat R_j)}{2},
\end{equation}
where $C$, $L$, and $R$ denote the centerline and boundaries, $d_F$ and $d_C$
are Fr\'echet and symmetric Chamfer distance, and
$r_i=\max(0.5,1-0.005e_i)$ with $e_i$ the ground-truth lane's minimum 3-D
distance to ego. Hungarian matching gives a one-to-one assignment, retaining
pairs with $d_{ij}<3.0$\,m.

For each accepted pair, centerline and boundaries are independently resampled
to 100 points, height is discarded, and their three symmetric XY Chamfer
distances are averaged. The frame score is the mean over accepted pairs, and
failure is defined as a score above $0.5\,\mathrm{m}$. Frames with no accepted match
are excluded.

\subsection{Results}
\label{sec:lanesegnet_results}

Table~\ref{tab:lanesegnet_results} compares all LaneSegNet monitors on the
4,439 validation frames with finite matched-lane Chamfer. Across this
population, mean matched-lane Chamfer is $0.50$\,m and mean $AP_{ls}^{*}$ is
$0.35$. Coverage denotes the fraction of frames classified as high-risk.

\noindent\textbf{1. Representation novelty identifies degraded geometry:}
Both image and BEV-based novelty provide useful failure ranking
(Table~\ref{tab:lanesegnet_results}). BEV spatial $2{\times}2$ is the stronger
density monitor. At its frozen operating point, the retained subset has mean
Chamfer $0.472$\,m and mean $AP_{ls}^{*}$ $0.386$, corresponding to 5.6\% lower
geometric error and 10.3\% higher $AP_{ls}^{*}$ than the full validation set.
Thus, atypical latent representations are associated with degraded map
geometry and can be used to selectively identify lower-quality predictions.

\noindent\textbf{2. Supervised latent monitoring extracts stronger
failure-relevant information:}
Supervised image and BEV monitors both improve substantially over their
density-based counterparts. Image--BEV fusion gives the strongest latent-only
point estimate (AUROC $0.780$, AP $0.761$), although its advantage over the
BEV-only supervised monitor is not statistically resolved. These results show
that frozen internal representations contain substantial information about
frame-level geometric failure beyond that captured by generic representation
novelty.

\noindent\textbf{3. Prediction outputs from the model:}
We consider two representations of LaneSegNet's generated set of lanes,
preserving the explicit geometric structure that is central to lane-segment
prediction~\cite{abraham2026georeformer}.
The first is a 167-D engineered descriptor computed entirely from predictions,
including confidence statistics, lane length, curvature, smoothness,
boundary-width consistency, boundary semantics, coarse spatial occupancy, and
prediction-to-prediction duplicate-lane distances. The second is a
permutation-invariant DeepSets-style encoder over the 64 highest-confidence
nearby lanes, using their centerline and boundary geometry and confidence,
with the engineered descriptor as global context. This gives a
structured-output alternative to the hand-engineered representation.

\noindent\textbf{4. Latent augmentation provides no resolved incremental benefit:}
Directly appending the tested image and BEV latent features to the engineered
prediction-output representation does not improve either AP or AUROC. Under
2,000 source-group bootstrap replicates, the changes are $-0.0135$ in AUROC
(95\% CI $[-0.0347,\,0.0080]$) and $-0.0093$ in AP
($[-0.0308,\,0.0123]$). Across five matched training seeds, the corresponding
hierarchical-bootstrap differences are $+0.0061$
($[-0.0158,\,0.0273]$) and $+0.0037$ ($[-0.0152,\,0.0243]$).

We additionally test leakage-safe score-level stacking of the latent and
engineered-output monitors. The stacker assigns nonzero weight to both but
improves AUROC by only $+0.0069$ (95\% CI $[-0.0021,\,0.0151]$) and AP by
$+0.0081$ ($[-0.0019,\,0.0179]$). Its selective effect is likewise small:
at 20\% rejection, retained failure rate changes from 32.86\% to 32.78\%,
and at 40\% from 23.47\% to 22.87\%. Thus, across both feature-level
concatenation and score-level fusion, the tested latent access provides no
statistically resolved or operationally meaningful incremental benefit over
the prediction-output monitor.

\noindent\textbf{5. The latent signal is robust to the failure threshold:}
Without retraining, we reevaluate the frozen monitor scores at Chamfer
thresholds from $0.3\,\mathrm{m}$ to $0.75\,\mathrm{m}$. Despite endpoint prevalence falling from
$0.804$ to $0.138$, image--BEV fusion maintains AUROC $0.703$--$0.781$ and
the BEV density monitor $0.653$--$0.737$; fusion AP remains above endpoint
prevalence at every threshold. Thus the latent failure signal is not specific
to the selected $0.5\,m$ boundary.

\begin{table*}[t]
\centering
\caption{\textbf{VAD monitoring across alternative planning endpoints.}
Each row specifies the information available to the monitor and the approach
used to convert it into a risk score. ``Latent + non-latent'' concatenates the planning representation with
ego state, predicted trajectory, and/or driving command. Results are AUROC and AP
on official validation.}
\label{tab:vad_endpoints}
\footnotesize
\setlength{\tabcolsep}{3.5pt}

\begin{tabularx}{\textwidth}{@{}p{2.25cm} X l c c@{}}
\toprule
Endpoint & Information available & Approach & AUROC & AP \\
\midrule

\multirow{5}{2.25cm}{ADE $>2$ m or agent collision}
& Four-level front-camera FPN (latent)
& PCA-128 + GMM NLL & .500 & .312 \\

& BEV (latent)
& PCA-128 + GMM NLL & .526 & .320 \\

& Ego state + command (non-latent)
& Supervised MLP & .753 & .546 \\

& Predicted trajectory + command (non-latent)
& Supervised MLP & .815 & .666 \\

& Planning representation (latent )
& Supervised MLP & .844 & .712 \\

& Ego state + trajectory + command (non-latent)
& Supervised MLP & \textbf{.900} & \textbf{.839} \\

& Planning representation + ego state + trajectory + command
  (latent + non-latent)
& Supervised MLP & .890 & .823 \\

 
\midrule

\multirow{3}{2.25cm}{ADE $>2$ m}
 & Planning representation (latent)
 & Supervised MLP & .868 & .724 \\
 & Ego + trajectory + command (non-latent)
 & Supervised MLP & \textbf{.924} & \textbf{.855} \\
 & Planning representation + ego + trajectory + command (latent + non-latent)
 & Supervised MLP & .917 & .839 \\
\midrule

\multirow{3}{2.25cm}{Cross-track $>1$ m}
 & Planning representation (latent)
 & Supervised MLP & .875 & .763 \\
 & Ego + trajectory + command (non-latent)
 & Supervised MLP & \textbf{.909} & \textbf{.827} \\
 & Planning representation + ego + trajectory + command (latent + non-latent)
 & Supervised MLP & .903 & .819 \\
\midrule

\multirow{4}{2.25cm}{Terminal heading $>10^\circ$}
 & Command only (non-latent)
 & Supervised MLP & .795 & .459 \\
 & Planning representation (latent)
 & Supervised MLP & .910 & .622 \\
 & Ego + trajectory + command (non-latent)
 & Supervised MLP & .922 & \textbf{.691} \\
 & Planning representation + ego + trajectory + command (latent + non-latent)
 & Supervised MLP & \textbf{.925} & .685 \\
\midrule

\multirow{3}{2.25cm}{Agent clearance $<1$ m}
 & Planning representation (latent)
 & Supervised MLP & .705 & .226 \\
 & Top-32 agent tokens + planner context (latent)
 & Agent-set attention & .697 & .219 \\
 & Ego + trajectory + command (non-latent)
 & Supervised MLP & .703 & \textbf{.229} \\
\midrule

\multirow{3}{2.25cm}{Boundary contact}
 & Planning representation (latent)
 & Supervised MLP & .780 & .365 \\
 & Ego + trajectory + command (non-latent)
 & Supervised MLP & \textbf{.818} & \textbf{.454} \\
 & Planning representation + ego + trajectory + command (latent + non-latent)
 & Supervised MLP & .802 & .422 \\
\bottomrule
\end{tabularx}
\end{table*}

\section{Monitoring End-to-End Planning with VAD}
\label{sec:vad}

We next apply the audit of Sec.~\ref{sec:evaluation_protocol} to end-to-end planning with
VAD~\cite{jiang2023vad} on nuScenes~\cite{caesar2020nuscenes}. VAD is useful for this comparison because
the same planning behavior can be monitored from internal image, BEV, and
planning representations, or from non-latent ego state and the emitted
trajectory.

\subsection{Representations}
\label{vad_representations}

\noindent\textbf{Image, BEV, and planning latents:}
For image-space analysis, we globally average each channel of all four
front-camera FPN levels. The resulting four 256-D summaries are concatenated
to form a 1,024-D multiscale image representation. For BEV analysis, we
globally average VAD's BEV feature tensor over its spatial dimensions,
yielding a 256-D representation. The planning latent is VAD's
frozen 512-D planning representation, formed by concatenating its 256-D
ego-agent and 256-D ego-map planning features. For the agent-clearance
analysis, we additionally consider the 32 highest-confidence agent tokens
with planner context.

\noindent\textbf{Non-latent state and outputs:}
The non-latent inputs are the 3-D driving command, a 9-D extracted ego-state
vector, and the command-selected predicted trajectory of six 2-D future
waypoints. Their concatenation forms the 24-D non-latent baseline. Appending
the 512-D planning latent produces a 536-D latent + non-latent representation for
the direct conditional audit.

\subsection{Monitoring Approaches}
\label{vad_methodology}
\noindent\textbf{Density-based representation novelty:}
We follow the density-monitoring protocol of
Sec.~\ref{sec:lanesegnet_methodology}, splitting training scenes into a 90\%
density-fit partition and a disjoint 10\% calibration partition. Image and
BEV features are standardized, reduced by unwhitened PCA to 128 dimensions,
and modeled with a diagonal-covariance GMM whose component count is selected
between 32 and 64 by training BIC. Negative log likelihood is used as the
continuous representation-novelty score. Youden's $J$ is used to select operating thresholds and official validation remains untouched
until evaluation.

\noindent\textbf{Supervised latent monitoring:}
Supervised VAD monitors use a two-hidden-layer
$128\rightarrow64$ MLP with inputs standardized from training
data only. The ADE-or-collision, ADE-only, cross-track, and terminal-heading monitors are trained on their respective endpoints. Agent clearance
additionally evaluates an ego-conditioned attention model over the 32
highest-confidence agent tokens with planner context.

\noindent\textbf{Model-output monitoring:}
The non-latent monitor uses VAD's command-selected predicted trajectory
together with ego state and driving command. The predicted trajectory contains
six future 2-D waypoints over the three-second horizon. We train the same supervised MLP on this non-latent input and
compare it with latent-only monitors.

To test incremental latent value, we additionally append the relevant latent
representation to the same non-latent baseline and evaluate the two models on
identical samples. 

\subsection{Failure Definitions and Results}
\label{vad_endpoints}

\begin{figure*}[t]
\centering
\includegraphics[width=\textwidth]{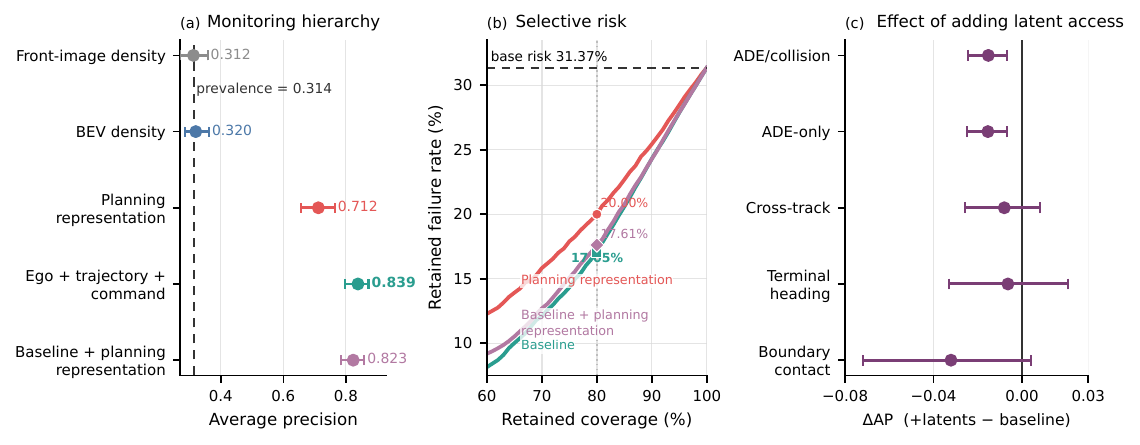}
\caption{\textbf{VAD monitoring hierarchy and conditional latent-value audit.}
\textbf{(a)} Average precision for front-image and BEV representation-novelty scores remain near
endpoint prevalence (0.314), while a supervised planning representation monitor is strongly
predictive. A non-latent monitor using ego state, predicted trajectory, and
driving command is stronger still, and appending the planning representation does not
improve it.
\textbf{(b)} Selective-risk curves. At 80\% retained
coverage, failure rate falls from 31.37\% overall to 20.00\% with the
planning representation monitor, 17.05\% with the non-latent baseline, and 17.61\% after
appending the planning representation.
\textbf{(c)} Paired change in average precision from appending the planning representation
to the complete non-latent baseline across endpoints. Error bars denote 95\%
scene-bootstrap confidence.}
\label{fig:vad_monitor_audit}
\end{figure*}

Table~\ref{tab:vad_endpoints} summarizes the monitoring results across all
failure definitions, while Fig.~\ref{fig:vad_monitor_audit} highlights the
main comparison between latent and non-latent monitoring. 

\noindent\textbf{1. ADE $>2$ m or agent collision:}
A frame with a complete three-second horizon is positive when either mean
displacement between predicted and expert trajectories over six future
waypoints exceeds $2$\,m, or the predicted trajectory introduces an agent
collision absent from the expert trajectory. The endpoint contains 1,606
positives among 5,119 validation frames (31.37\%). Image and BEV-based novelty are nearly
uninformative, whereas the supervised planning representation monitor is strongly
predictive. The complete non-latent baseline is stronger still, but appending
the planning representation decreases AUROC.

\noindent\textbf{2. Mean ADE $>2$ m:}
ADE is the mean Euclidean displacement between predicted and expert
trajectories over the six future waypoints. It measures demonstration
disagreement but mixes lateral path error with longitudinal speed or timing
differences. Training the monitors directly on ADE shows that the non-latent baseline is strongest, while appending the planning representation is numerically worse. The decomposition motivates more
specific geometric endpoints: at three seconds, 79.2\% of eligible frames
exceed $1$\,m along-track error, compared with 31.5\% for the identically
thresholded cross-track error.

\noindent\textbf{3. Cross-track disagreement $>1$ m:}
To isolate lateral disagreement, we express error in the
local expert-path frame and project it onto the path normal. A frame is
positive when absolute cross-track error at three seconds exceeds $1$\,m;
near-stationary records are excluded. The
non-latent baseline again outperforms the planning representation monitor, with no resolved
benefit from latent augmentation. Cross-track disagreement
also becomes substantially harder to monitor on curved paths: prevalence
rises from 24.7\% on nearly straight paths to 52.5\% on strongly curved paths,
where the best AUROC falls to approximately $0.70$.

\noindent\textbf{4. Terminal-heading disagreement $>10^\circ$:}
Cross-track error measures lateral position but not whether the predicted
trajectory is heading in the correct direction. We therefore compare the
direction of motion over the final second (2--3\,s) of the predicted and
expert trajectories and flag frames whose headings differ by more than
$10^\circ$. The one-second interval reduces sensitivity to individual
waypoint noise, and the $10^\circ$ threshold provides a nontrivial endpoint
without making small angular deviations failures. This analysis reveals an important confound rather than a new monitoring
success. Heading disagreement is rare for straight driving but common during
turns, so even driving command alone reaches AUROC $0.795$. Within strongly
curved trajectories, where heading failures are most common, all monitors are
near chance. The high pooled AUROCs therefore largely reflect recognition of
turning versus straight-driving regimes rather than reliable discrimination
of failures within the same regime; the small AUROC increase from latent
augmentation is not statistically resolved.

The preceding endpoints are all defined relative to the expert
trajectory. Because the non-latent baseline includes VAD's predicted
trajectory, strong performance could partly reflect
predictability of expert--planner disagreement from the planner output itself.
We therefore also evaluate two physical endpoints whose labels depend on
scene geometry unavailable to the non-latent monitor.

\noindent\textbf{5. Ground-truth-agent clearance $<1$ m:}
We compute the
minimum distance between VAD's planned ego footprint and future oriented
ground-truth vehicle and pedestrian boxes over the six future timestamps.
Ground-truth agent geometry is used only to construct the label and is never
available to a monitor. The endpoint occurs in 12.78\% of validation frames.
Predictability is modest, and the planning representation does not significantly outperform the non-latent
baseline.

\noindent\textbf{6. Planner-introduced boundary contact:}
We transform predicted and expert ego footprints into nuScenes map
coordinates and test them against the boundary of the union of road-segment
and lane polygons. A frame is positive when the predicted footprint contacts
the boundary at a timestamp for which the expert footprint does not. The
endpoint occurs in 15.26\% of validation frames. Although the defining map
geometry is unavailable to the non-latent monitor, the complete non-latent
baseline remains stronger than the planning representation monitor, and appending the planning representation provides no statistically resolved improvement.

Figure~\ref{fig:vad_monitor_audit} summarizes the resulting audit: latent
representations can be strongly predictive of failure, but across the tested
comparisons direct latent access provides no resolved improvement over a
strong non-latent monitor of ego state and model outputs.

\section{Discussion and Limitations}
\label{sec:discussion}

\subsection{What Does Latent Predictability Establish?}
\label{sec:discussion_interpretation}

Our results separate three claims that are easy to conflate in failure-monitor
evaluation. First, representation novelty may correlate with model error.
Second, a supervised monitor may predict failure from an internal
representation even when generic novelty does not. Third, and more strongly,
direct access to that representation may provide predictive information beyond
what can already be recovered from non-latent model state and outputs.

The first two claims hold strongly in our experiments, but they do not imply
the third. In LaneSegNet, BEV novelty predicted by density tracks geometric degradation and
supervised latent probes improve substantially over density monitoring. However, prediction-output features outperform the latent probes and matched
latent augmentation yields no statistically resolved gain. In VAD, image and BEV space novelty is nearly uninformative, while a supervised
latent planning representation probe strongly predicts deviation from expert demonstration trajectories. Nevertheless,
ego state and predicted trajectory geometry are stronger jointly, and
appending the latent planning representation again provides no resolved improvement. Thus, strong
marginal latent predictability does not establish
incremental monitoring value.

For expert-trajectory-relative endpoints such as ADE and cross-track
disagreement, strong performance from a monitor that observes the generated
trajectory is unsurprising because that output is closely related to the
quantity defining the label. The physical endpoints provide a stricter test. For agent clearance, the planning-representation monitor performs similarly to the non-latent baseline; for boundary contact, direct latent augmentation provides no statistically resolved improvement.

\subsection{Implications for Failure-Monitor Evaluation}
\label{sec:discussion_protocol}

Our experiments suggest three practical requirements for evaluating
latent-monitor claims.

\noindent\textbf{Use strong non-latent baselines.}
Latent monitors should be compared against combinations of available
non-latent information rather than isolated low-information controls. In VAD,
ego state and predicted trajectory are each weaker than the planning-
representation monitor, while their combination is substantially stronger.
Evaluating only either component would therefore support a qualitatively
different conclusion.

\noindent\textbf{Test augmentation, not only competition.}
Comparing separately trained latent and non-latent monitors does not establish
whether they contain complementary predictive information. A more direct test
compares a baseline with the same baseline augmented by the latent
representation on identical evaluation samples using paired statistical
comparisons. This tests the additional predictive value of latent access under
the chosen estimator.

\noindent\textbf{Stress-test the failure definition.}
Monitor performance can change substantially with the endpoint. Trajectory
disagreement, lateral deviation, agent clearance, and boundary contact differ
in both information requirements and predictability. Endpoints whose labels
depend on information hidden from the baseline are particularly useful for
testing apparent output shortcuts. Regime-conditioned evaluation is also
important: VAD's terminal-heading endpoint obtains high pooled AUROC largely
because failure prevalence differs sharply across maneuver regimes, while
within-regime discrimination is much weaker.

\subsection{Scope and Limitations}
\label{sec:limitations}

\noindent\textbf{Models and representations}
We study two models and tasks rather than attempting to
establish a universal property of autonomous-driving representations. Additional architectures,
datasets, and distribution shifts are needed to determine how broadly these
results generalize.

\noindent\textbf{Monitor capability}
Our supervised monitors are intentionally lightweight probes over frozen representations. Beyond direct concatenation, we also tested a leakage-safe out-of-fold residual stacker that fits a latent head to the output monitor's errors; it assigned nonzero weight to the latent score yet produced no resolved AUROC or AP improvement, indicating the null is not an artifact of a single naive fusion. More expressive interaction or late-fusion architectures could still test a different hypothesis, and we do not claim to have exhausted the space of monitors.

\noindent\textbf{Failure endpoints}
LaneSegNet's Chamfer endpoint conditions on an accepted lane match and
therefore does not capture missed or extra lane predictions. The
VAD experiments are open-loop: agent clearance, boundary contact, and
corrected collision remain proxies for closed-loop safety. 

Within this empirical scope, the central evaluation distinction is consistent:
strong latent-only failure prediction does not by itself establish that
\emph{direct latent access} provides additional monitoring value unless it is
tested conditionally against a strong non-latent baseline.

\section{Conclusion} \label{sec:conclusion}

We developed and audited inference-time failure monitors for two distinct autonomous-driving tasks: online vectorized map generation with LaneSegNet and end-to-end planning with VAD. We monitored multiple information sources across the two models, including image and BEV features, internal planning representations, and model outputs. We evaluated density-based and supervised monitors and compared them with strong non-latent baselines across multiple failure definitions.

For LaneSegNet, both representation novelty and supervised latent probes
predict high geometric-error frames. A monitor using only the model's
predicted lane outputs is stronger still, but adding latent features to this
baseline provides no statistically resolved improvement. To our knowledge,
this is the first post-hoc frame-level failure monitor for online vectorized
map generation.

For VAD, novelty in image and BEV space is largely uninformative, while
a supervised probe over the 512-D planning representation strongly predicts
expert-trajectory disagreement. A non-latent monitor using ego state and the
predicted trajectory is stronger, and directly adding the planning
representation again provides no resolved improvement. The broader pattern
persists across alternative failure definitions, including physical
agent-clearance and boundary-contact endpoints whose defining scene geometry
is unavailable to the non-latent monitor.

Across both tasks, a latent representation can predict failure well without
direct access to it improving a strong non-latent monitor. Latent-monitor
evaluations should therefore include strong state and output baselines,
directly test whether latent access improves those baselines, and verify
conclusions across failure definitions. The key question is not simply
whether a latent predicts failure, but whether giving the monitor direct
access to it improves failure prediction.

{
    \small
    \bibliographystyle{ieeenat_fullname}
    \bibliography{main}
}

\end{document}